\documentclass[runningheads]{llncs}
\usepackage[T1]{fontenc}
\usepackage{graphicx}
\usepackage{amsmath}
\usepackage{amssymb}
\usepackage{xcolor}
\usepackage{censor}

\usepackage[colorlinks=true, linkcolor=black, urlcolor=blue, citecolor=black]{hyperref}
\begin{document}
\title{UPCMR: A Universal Prompt-guided Model for Random Sampling Cardiac MRI Reconstruction}


%
\titlerunning{UPCMR: A Universal Prompt-guided Model}
%

\author{Donghang Lyu\inst{1} \and Chinmay Rao\inst{1} \and Marius Staring\inst{1} \and Matthias J.P. van Osch\inst{1} \and Mariya Doneva\inst{4} \and Hildo J. Lamb\inst{1} \and Nicola Pezzotti\inst{2,3}}

\institute{Department of Radiology, Leiden University Medical Center, Leiden, The Netherlands\\ 
\email{d.lyu@lumc.nl}
\and
Cardiologs, Philips, Paris, France
\and
Faculty of Computer Science, Eindhoven University of Technology, Eindhoven, The Netherlands
\and
Philips Innovative Technologies, Hamburg, Germany}

%
%

\authorrunning{D. Lyu et al.}

\maketitle              
\begin{abstract}
Cardiac magnetic resonance imaging (CMR) is vital for diagnosing heart diseases, but long scan time remains a major drawback. To address this, accelerated imaging techniques have been introduced by undersampling k-space, which reduces the quality of the resulting images. Recent deep learning advancements aim to speed up scanning while preserving quality, but adapting to various sampling modes and undersampling factors remains challenging. Therefore, building a universal model is a promising direction. In this work, we introduce UPCMR, a universal unrolled model designed for CMR reconstruction. This model incorporates two kinds of learnable prompts, undersampling-specific prompt and spatial-specific prompt, and integrates them with a UNet structure in each block. Overall, by using the CMRxRecon2024 challenge dataset for training and validation, the UPCMR model highly enhances reconstructed image quality across all random sampling scenarios through an effective training strategy compared to some traditional methods, demonstrating strong adaptability potential for this task.

\end{abstract}

%
%
%
\section{Introduction}
Cardiac Magnetic Resonance Imaging often requires multiple breath-holds for a comprehensive heart scan, leading to potential patient discomfort and slice misalignment. Moreover, the combination of data from multiple cardiac cycles can be problematic, especially in cases of arrhythmia, limiting real-time imaging to lower spatio-temporal resolutions. To address these challenges, undersampling techniques have been explored over the years to accelerate the whole scanning process. In recent years, deep learning methods~\cite{qin2018convolutional,schlemper2017deep,sriram2020end} have gained significant attention and are increasingly being used for MRI reconstruction, aiming to restore image details effectively, even with high acceleration factor. However, managing various undersampling scenarios with a single model remains challenging. The main difficulty lies in achieving and maintaining high adaptability across these varied conditions. Inspired by the rapid development of prompt learning~\cite{visp,mio,xin2023fill}, we introduce two kinds of learnable prompts, aiming to improve the adaptability across different contrasts, k-space trajectories, and acceleration factors.

Prompt learning adapts large pre-trained models, also known as foundation models, to new tasks by providing specific hints. It has been widely used in the natural language processing (NLP)~\cite{wei2022chain,zhang2022automatic} and computer vision (CV)~\cite{kirillov2023segment,potlapalli2024promptir,xin2023fill} tasks. In the medical imaging field, prompts have been extensively utilized for segmentation tasks. Medical Segment Anything Model (MedSAM)~\cite{ma2024segment} uses box prompt to provide spatial hints, enabling accurate segmentation of masks within the specified box. Uniseg~\cite{uniseg} introduces a universal prompt, which generates corresponding task-specific prompts and improves the correlation among different tasks. For MRI reconstruction, PromptMR~\cite{xin2023fill} employs learnable prompts to adapt itself to two different tasks: cine MRI reconstruction and T1-weighted (T1w) and T2-weighted (T2w) MRI reconstruction. Inspired by the general strategy of prompt-based guidance, we proposed an unrolled all-in-one model called UPCMR, which integrates learnable prompts into the UNet structure in each block so as to better adapt to different undersampling scenarios. In addition to model design, an effective training strategy is crucial for handling versatile random sampling. Inspired by the widespread use of curriculum learning~\cite{cl} in similar tasks, we also explored its effectiveness with two formats and selected the most effective one as the final training strategy for the UPCMR method.

\section{Preliminaries and Dataset}
\subsection{Preliminaries}
Consider reconstructing a sequence of complex-valued MR images $x$ $\in$ $\mathbb{C}^{T \times H \times W}$ from multi-coil undersampled k-space measurements $y$, where $H$ and $W$ denote the height and width of each frame, respectively, $T$ represents the sequence length. Therefore, the goal is to reconstruct $x$ from $y$, formulated as
\begin{equation} 
    \operatorname*{argmin}_{x} \| y - A x \|_2^2 + \lambda \mathcal{R}(x),
\end{equation}
where $A$ is the linear forward operator composed of coil sensitivity encoding $S$, 2D Fourier transform $\mathcal{F}$, and undersampling mask $M$, $\mathcal{R}$ represents the regularisation terms with $\lambda$ as a hyper-parameter controlling the regularization strength. For deep unrolled methods, $\mathcal{R}$ represents a trainable neural network block. Some previous deep unrolled models based on the ADMM optimization algorithm such as Deep-ADMM net~\cite{admm} introduce an intermediate variable $z$, also computed via a learnable block. When constraining $z$ to be equal to $x$, the above problem is reformulated as
\begin{equation}
    \operatorname*{argmin}_{x,z}  \| y - A x \|_2^2+ \mu \| x -z \|_2^2+ \lambda \mathcal{R}(z).
\end{equation}
The unrolled model thus learns a sequence of transition where $x^{i}$ passes through a neural network block to produce $z^{i}$, and $z^{i}$ is used to generate $x^{i+1}$ through the data consistency (DC) layer. These operations are represented as
\begin{equation}
    z^{i} = f_\theta^i(x^i),
\end{equation}
\begin{equation}
    x^{i+1} = DC(z^{i},y,\lambda_{0},\Omega)=A^{\dagger} \Lambda Az^{i}+\frac{\lambda_{0}}{1+\lambda_{0}}A^{\dagger}y
\end{equation}
where $A^{\dagger}$ denotes the Hermitian operation of $A$, $\lambda_{0}$ is a regularization parameter and we make it approach infinity to ensure the preservation of sampled k-space information, $\Omega$ is an index set of the acquired k-space samples and $\Lambda$ is a diagonal matrix, whose diagonal values are
\begin{equation}
\Lambda_{kk} =
\left\{
\begin{array}{lcl}
& 1          & {if \  k \notin \Omega}\\
& \frac{1}{1+\lambda_{0}}          & {if \  k \in \Omega}\\
\end{array} \right.
\end{equation}
\subsection{CMRxRecon2024 Dataset}
The CMRxRecon2024 dataset~\cite{data2024} includes data from 330 healthy volunteers scanned with 3 Tesla magnets. Task 2 features multi-contrast k-space data (Cine, Aorta, Mapping, Tagging) with anatomical views such as long-axis, short-axis, LVOT, and aortic views (transversal and sagittal). Each contrast consists of 5-15 slices, segmented into 12-25 cardiac phases with a temporal resolution of ~50 ms. Key geometrical parameters are a spatial resolution of 1.5×1.5 mm², slice thickness of 8.0 mm, and a slice gap of 4.0 mm. The dataset includes three k-space trajectories (uniform, Gaussian, pseudo radial) with temporal/parametric interleaving, and various acceleration factors (4$\times$, 8$\times$, 12$\times$, 16$\times$, 20$\times$, 24$\times$) without including autocalibration signal (ACS) for calculation. For the ACS area, it includes the central 16 lines for uniform and Gaussian types, and the central 16$\times$16 regions for the pseudo radial type. Training data includes 200 subjects with fully-sampled k-space data across all combinations of k-space trajectories
and acceleration factors, while 60 validation subjects are provided with random undersampling settings. The ground-truth of validation set is unavailable, and performance is assessed by submitting the reconstructed central cropped parts of the images to the platform.
\section{Methodology}
\subsection{Model Structure}
\begin{figure}[tb]
\centering
\includegraphics[width=\textwidth]{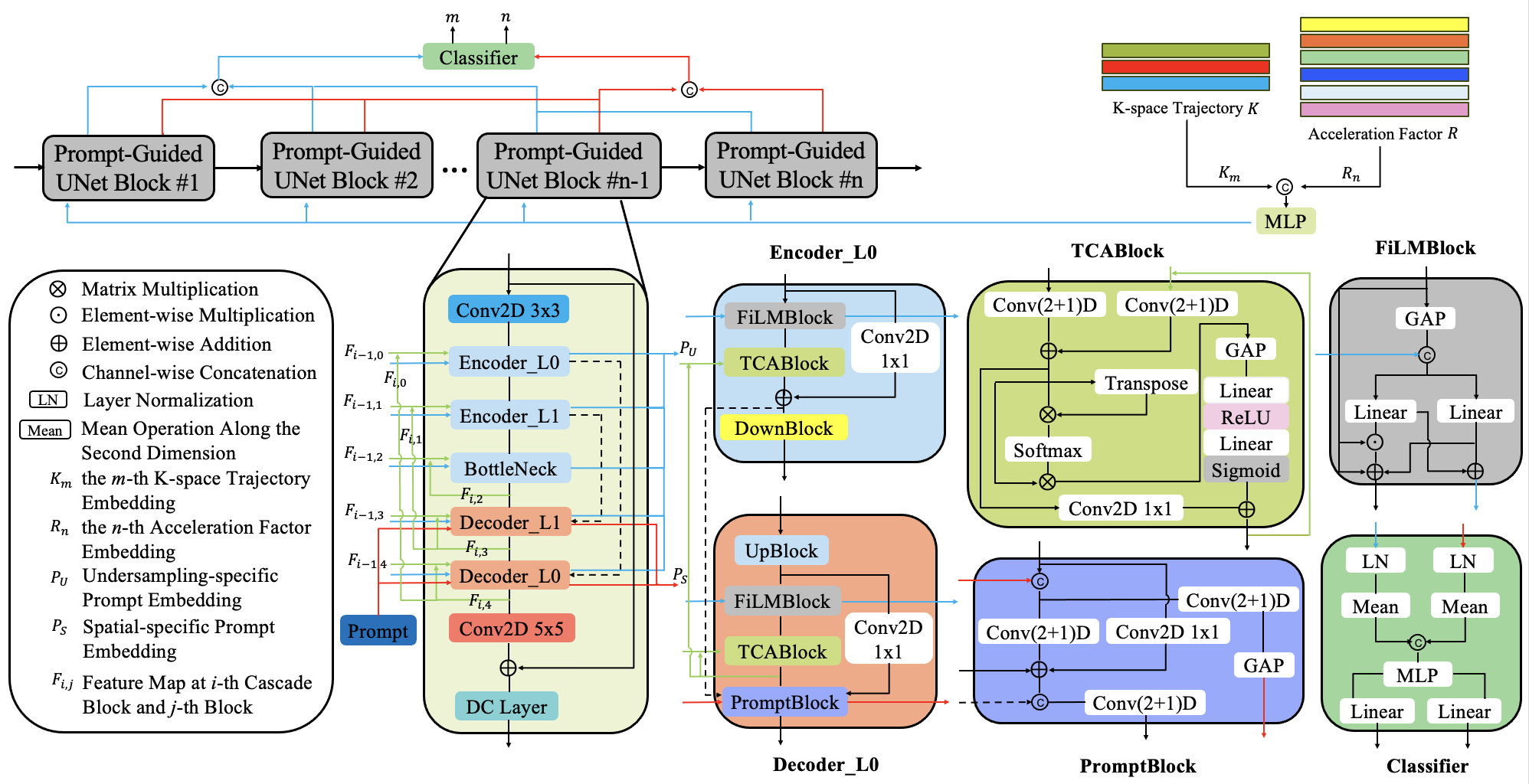}
\caption{Overview of the UPCMR model. In each cascade block, green lines represent feature maps transitioning from the previous one to the next block. Blue and red lines highlight the direction of undersampling-specific prompts and spatial-specific prompts, respectively. Black dashed lines refer to the skip connections between the corresponding encoder and decoder components.} \label{fig1}
\end{figure}

\textbf{UPCMR} We follow an unrolled design used by some previous methods~\cite{qin2018convolutional,xin2023fill}, as shown in Fig.~\ref{fig1}. 
From a global perspective, in addition to common inputs like undersampling k-space, corresponding mask, coil-combined image sequence with 2 channels, and coil sensitivity map, an undersampling-specific prompt embedding is also provided as input. Given that both k-space trajectory and acceleration factor are prior information, we build two learnable prompt pools: one for k-space trajectory $K \in \mathbb{R}^{3 \times C}$ and another for the acceleration factor $R \in \mathbb{R}^{6 \times C}$, where $C$ is the embedding length, which matches the channel number of the UNet block. When selecting the $m^{th}$ k-space trajectory $K_{m}$ and $n^{th}$ acceleration factor $R_{n}$, these prompt embeddings are combined by a multi-layer perceptron (MLP) and fed into each cascade block to introduce undersampling-related information, enhancing model adaptability. UPCMR outputs include the reconstructed image series, as well as predictions for k-space trajectory and acceleration factor classes. Here, an additional classification block is used during training. The undersampling-specific prompt embedding $P_{U}$ and the spatial-specific prompt embedding $P_{S}$ from all cascade blocks are concatenated to make classification so as to further condition the model.
\\
\\
\textbf{Prompt-guided UNet Block} Each block of the cascading structure incorporates a 2-level UNet featuring some extra Conv2D layers and a bottleneck block that mirrors the encoder block's structure. The channel number is fixed at 64 across all layers, with downsampling and upsampling operations only occur at the first level. In the encoder part, for the input feature map at $i^{th}$ UNet block and $j^{th}$ encoder block, $F_{i,j} \in \mathbb{R}^{B \times T \times C \times H \times W}$, it first passes through a FiLMBlock~\cite{film}, interacting with the undersampling-specific prompt $P_{U} \in \mathbb{R}^{B \times C}$, where $B$ denotes batch size. $F_{i,j}$ is processed by global average pooling (GAP) operation to combine with $P_{U}$ along the channel dimension. Then two separate linear layers generate weight embedding $W_{P} \in \mathbb{R}^{B \times T \times C}$ and bias embedding $B_{P} \in \mathbb{R}^{B \times T \times C}$. These embeddings are applied to $F_{i,j}$ through element-wise multiplication and addition to produce the output $F_{i,j+1}$ with prompt information. Additionally, $W_{P}$ and $B_{P}$ add together and do average along the time dimension to form an updated undersampling-specific prompt embedding that integrates image sequence information. The updated undersampling-specific prompt embedding from each block are concatenated to form the final $P_{U}$ of the current cascade block. After that, a temporal-channel attention block (TCABlock) is applied. Compared to traditional Conv3D or \textit{Conv(2+1)D}~\cite{conv21d} operations, the TCABlock incorporates a simple attention mechanism along temporal and spatial dimensions, aiming to better exploit the spatio-temporal correlation across the whole sequence. Furthermore, inspired by the CRNN-i operation from the CRNN-MRI~\cite{qin2018convolutional} model, the feature maps from the previous cascade block are added to strengthen cascade connections and retain some important features. As shown in Fig.~\ref{fig1}, two inputs, $F_{i,j}$ and $F_{i-1,j}$, pass through separate \textit{Conv(2+1)D} layer and add together to obtain an intermediate feature map $F_{tmp}$. Then it multiplies with the transposed $F_{tmp}$ to compute temporal correlations between frames and generate an attention map using the softmax function. This attention map is multiplied with $F_{tmp}$ to produce the final output, incorporating temporal attention information. Channel attention is applied following the same design as the SE block~\cite{se}. After the TCABlock, the feature map is processed through a \textit{Conv(2+1)D} layer, which either downsamples or maintains the current shape. Moreover, temporal circular padding~\cite{tcp} is used with all \textit{Conv(2+1)D} layers to account for the circular nature of cardiac MRI.

The decoder blocks follow a design similar to the encoder blocks but include a PromptBlock with a learnable spatial-specific prompt $P_{S} \in \mathbb{R}^{B \times 1 \times C \times H \times W}$, inspired by~\cite{potlapalli2024promptir,xin2023fill}. This addition helps the model adapt to different undersampling cases by enriching the spatial context. The PromptBlock's pipeline begins with concatenating $F_{i,j}$ with $P_{S}$ along the channel dimension. This is followed by two branches: one generates the updated spatial-specific prompt embedding $P_{S} \in \mathbb{R}^{B \times C}$, while the other one uses convolutional layers to combine with the corresponding encoder output, generating the feature map for subsequent layers. Similar to the operation for undersampling-specific prompt embeddings, the updated spatial-specific prompt embedding from each block are concatenated to form the final $P_{S}$ of the current cascade block.

\subsection{Loss Function}
Given the three outputs of UPCMR, the loss function comprises the classification loss $\mathcal{L}_{cls}$ and the reconstruction loss $\mathcal{L}_{rec}$. The $\mathcal{L}_{cls}$ is the sum of the cross-entropy losses for the k-space trajectory class and the acceleration factor class. The $\mathcal{L}_{rec}$ is the weighted sum of an L1 loss term and an SSIM loss term, defined as follows:
\begin{equation}
    \mathcal{L}_{rec} = \lambda_{l1} \| I_{rec} - I_{gnd} \|_1 + \lambda_{ssim}(1-SSIM(|I_{rec}|, |I_{gnd}|)),
\end{equation}
where $I_{rec}$ denotes the reconstructed CMR image sequence and $I_{gnd}$ represents the original ground-truth image sequence, both of which are double-channeled. For the SSIM loss calculation, we use the absolute value to compute the loss. Based on the empirical study from~\cite{loss}, we set $\lambda_{l1}$=0.16 and $\lambda_{ssim}$=0.84. Finally, the overall loss function is as follows:
\begin{equation}
    \mathcal{L} = \lambda_{cls}\mathcal{L}_{cls}+\mathcal{L}_{rec}
\end{equation}
Considering classification as a minor task relative to the main reconstruction task, we set a relatively small weight of $\lambda_{cls}$=0.025.
\subsection{Training Strategy}
For random undersampling, an efficient training strategy is essential. Given thousands of slices, each with 3$\times$6=18 different undersampling scenarios, iterating through all in each epoch is time-consuming and impractical under the situation of limited GPU resource. Therefore, our policy is to randomly pick a slice with a random k-space trajectory and acceleration factor for each training sample. This approach highly reduces training time and ensures the model can still handle diverse scenarios. 

Another challenge is the varying difficulty of reconstruction across different undersampling modes, where, for example, an acceleration factor of 24 is much more difficult than one of 4. Therefore, equally random sampling might not be an optimal strategy for the model training. Inspired by the concept of curriculum learning~\cite{cl} and its extensive applications~\cite{mio,cnlp}, we explore incorporating this method into our model training process. Firstly, we trained the model for 90 epochs, divided into four stages:
\begin{enumerate}
    \item Train the model for 10 epochs using \{\textit{uniform}\} k-space sampling and an acceleration factor of \{4\}.
    \item Train the model for 20 epochs using \{\textit{uniform, }\textit{gaussian}\} k-space with the sampling probabilities \{0.2, 0.8\} and acceleration factors \{4, 8, 12\} with the sampling probabilities \{0.04, 0.48, 0.48\}.
    \item Train the model for 30 epochs using \{\textit{uniform}, \textit{gaussian}, \textit{pseudo\_radial}\} k-space with the sampling probabilities \{0.1, 0.1, 0.8\} and acceleration factors \{4, 8, 12, 16, 20, 24\} with the sampling probabilities \{0.02, 0.02, 0.03, 0.31, 0.31, 0.31\}.
    \item Train the model for 30 epochs using all the k-space and all the acceleration factors with the equal sampling probability.
\end{enumerate}
In the above 4 stages, we start with easier cases to help the model grasp basic reconstruction principles and adapt to less noisy data. Gradually, we introduce more complex k-space trajectories and higher acceleration factors to increase difficulty. The final stage involves combined training, which further boosts the model's adaptability.

Although the above training strategy aligns with curriculum learning principles, using the same full model structure and expecting it to handle multiple acceleration factors may not be optimal. This approach increases the training difficulty of the UPCMR model, making it harder to learn and differentiate between these internal variations. Moreover, assigning the optimal sampling probabilities at each stage can be challenging, which may hinder the model's overall performance. Therefore, we propose an alternative training strategy applying curriculum learning in a sequential way, divided into 7 stages. In each stage, all three k-space trajectories are equally likely to be included, with a total of 50 training epochs. The key differences across stages lie in the number of cascade blocks in the UPCMR model and the introduction of acceleration factors, as detailed below:
\begin{enumerate}
    \item Initialize the model with 3 cascade blocks and train for 50 epochs with an acceleration factor of \{4\}.
    \item Add a new block to the model (total 4 blocks), train for 40 epochs with an acceleration factor of \{8\}, followed by 10 epochs using acceleration factors \{4, 8\}, with the sampling probabilities of \{1/2, 1/2\}.
    \item Add a new block to the model (total 5 blocks), train for 40 epochs with an acceleration factor of \{12\}, followed by 10 epochs using acceleration factors \{4, 8, 12\}, with the sampling probabilities of \{1/3, 1/3, 1/3\}.
    \item Add a new block to the model (total 6 blocks), train for 40 epochs with an acceleration factor of \{16\}, followed by 10 epochs using acceleration factors \{4, 8, 12, 16\}, with the sampling probabilities of \{1/4, 1/4, 1/4, 1/4\}.
    \item Add a new block to the model (total 7 blocks), train for 40 epochs with an acceleration factor of \{20\}, followed by 10 epochs using acceleration factors \{4, 8, 12, 16, 20\}, with the sampling probabilities of \{1/5, 1/5, 1/5, 1/5, 1/5\}.
    \item Add a new block to the model (total 8 blocks), train for 40 epochs with an acceleration factor of \{24\}, followed by 10 epochs using acceleration factors \{4, 8, 12, 16, 20, 24\}, with the sampling probabilities of \{1/6, 1/6, 1/6, 1/6, 1/6, 1/6\}.
    \item  Train the complete model for 50 epochs using all the k-space trajectories and all the acceleration factors with the equal sampling probability.
\end{enumerate}
In the first 6 stages, the model learns one acceleration factor at a time in the first 40 epochs, followed by 10 epochs of fine-tuning with all learned acceleration factors for comprehensive adjustment. Within each stage, a new cascade block is added to align with the increasing difficulty of the acceleration factor. Additionally, initializing the previous blocks with the trained weights from the last stage, we ensure smoother progression and better model adaptation. In the final stage, the model is trained with all possible scenarios to further refine its performance and ensure robust adaptability. In general, this training strategy follows a sequential approach with respect to model structure and acceleration factors, gradually increasing the model's complexity and training samples' difficulty to align with the principles of curriculum learning, helping the model solidify its understanding of internal differences between various undersampling scenarios.
\section{Experiments}
\subsection{Implementation Details}
During training, we generated the coil sensitivity map (CSM) from the time-averaged autocalibration signals (ACS) using the ESPIRIT algorithm~\cite{espirit}. The training data was normalized by first transforming the k-space data to the image domain with the Inverse Fast Fourier Transform (IFFT), normalizing the multi-coil images by their maximum absolute value, and then transforming the data back to k-space domain with the Fast Fourier Transform (FFT). Additionally, to further stabilize the model training, we incorporated z-score normalization and un-normalization at the start and end of each block in the cascade.

The models were implemented in Pytorch and trained on Nvidia A100 GPU with 80GB of GPU memory. To optimize GPU memory usage, we utilized mixed precision training~\cite{mixp}, enabling us to use a UPCMR model with up to 8 blocks. In the first training strategy, we used AdamW as the optimizer with an initial learning rate of 2e-4 and a weight decay of 1e-3. The learning rate was reduced by a factor of 0.8 every 10 epochs. The second training strategy also used AdamW with an initial learning rate of 2e-4 and a weight decay of 1e-2 but introduced a cosine scheduler with a 6-epoch warm-up, and the minimal learning rate was set to 2e-5. In the final stage, the initial learning rate was 1e-4, reduced by a factor of 0.8 every 5 epochs, with a minimum of 8e-6, dropping to 1e-6 in the last epoch. The batch size was set to 1. During training, we split 1380 samples for training and 16 samples for validation. Each validation sample was subjected to 18 different undersampling scenarios and each slice to increase the number of validation samples and better measure the model's overall performance. For evaluation, peak signal-to-noise ratio (PSNR), structural similarity (SSIM) and normalized mean squared error (NMSE) were used as image quality metrics, focusing on the cropped central region of each validation case.

\subsection{Results}

\begin{table}
\centering
\caption{Results of the methods on the central cropped area of validation set from the leaderboard. $\text{CL}_{1}$ refers to the first training strategy, while $\text{CL}_{2}$ is the second one. The best result is highlighted in bold.}\label{tab1}
\begin{tabular}{l|l|l|l}
\hline
Method &  PSNR $\uparrow$ & SSIM $\uparrow$ & NMSE $\downarrow $\\
\hline
ZF &  19.4182 & 0.4610 & 0.2853\\
SENSE &  24.4780 & 0.6200 & 0.1734\\
GRAPPA & 24.9678 & 0.6449 & 0.1554\\
\hline
 UPCMR w/o $\text{CL}$ & 25.3694 & 0.6854 & 0.1422\\
 UPCMR-$\text{CL}_1$ w/o FiLMBlock & 25.2132 & 0.6755 &0.1484\\
 UPCMR-$\text{CL}_1$ w/o TCABlock & 25.0871 & 0.6718 &0.1487\\
 UPCMR-$\text{CL}_1$ w/o PromptBlock & 25.4463 & 0.6858 &0.1409\\
 UPCMR-$\text{CL}_{1}$ & 25.6070 & 0.6904 & 0.1391 \\
 UPCMR-$\text{CL}_{2}$ & \textbf{26.6361} & \textbf{0.7338} &\textbf{0.1036}\\
 \hline
\end{tabular}
\end{table}
\begin{figure}[tbh!]
\centering
\includegraphics[width=\textwidth]{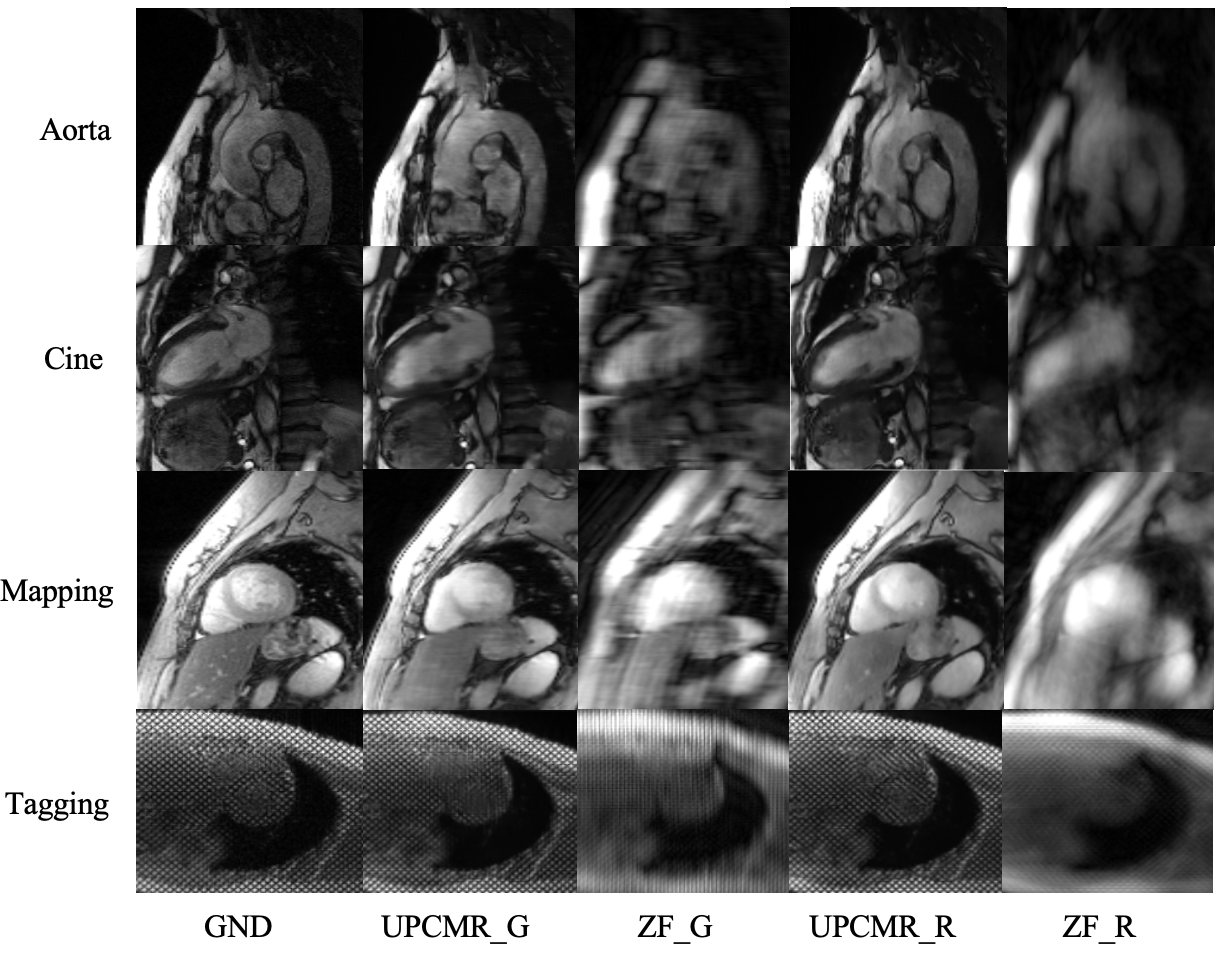}
\caption{Visualization of UPCMR reconstruction results for four contrasts under two k-space trajectories with an acceleration factor of 20. 'GND' indicates images from fully-sampled k-space, 'ZF' denotes images from zero-filled undersampled k-space, 'G' refers to Gaussian k-space, and 'R' stands for pseudo radial k-space. Images are cropped to focus on the central cardiac region. Rows from top to bottom show Aorta (sagittal), Cine (LAX), Mapping (T1), and Tagging.} \label{fig2}
\end{figure}
We compare the proposed model with two traditional methods, SENSE and GRAPPA, on the given validation set. Some ablation studies evaluate (1) the effectiveness of curriculum learning versus a combined learning strategy, where equal sampling probabilities are assigned to each combination of k-space trajectories and acceleration factors in the whole training procedure, (2) the impact of key components, FiLMBlock, TCABlock, and PromptBlock, on UPCMR's performance under the first curriculum learning strategy, TCABlock is replaced with a Conv3D layer to assess its effect, and (3) the relative performance of two different curriculum learning strategies. Results are summarized in Table~\ref{tab1}. Then Fig.~\ref{fig2} visualizes the reconstruction performance of UPCMR using the second curriculum learning strategy under some challenging undersampling scenarios from the splitting validation set.

\section{Discussion and Conclusion}


Table~\ref{tab1} shows that each key component block enhances performance, and the curriculum learning strategy outperforms the combined learning approach. Moreover, the second curriculum training strategy has been proven more effective, emphasizing the importance of learning each acceleration factor gradually and making the model size align with it. However, as shown in Fig.~\ref{fig2}, UPCMR struggles with high acceleration factors, resulting in detail loss and blurriness in the cardiac region, indicating room for improvement. For the second curriculum learning strategy, due to its time-consuming nature, we only set 50 training epochs for each stage. Consequently, the model did not fully converge at each stage, limiting the upper bound of final reconstruction performance. Training the UPCMR model with more epochs at each stage could further improve the final performance. Additionally, the UPCMR model may have some design drawbacks. While TCABlock is intended to capture spatio-temporal correlations across the entire sequence, its simple structure and operation might not effectively exploit these features, potentially limiting performance, especially at higher acceleration factors. To further enhance the model, incorporating techniques like convolutional recurrent operations could be considered. Additionally, there is a significant difference between contrasts, which can complicate the reconstruction process. The current UPCMR model doesn't consider about it, and introducing contrast-specific prompts might further enhance the reconstruction performance for each contrast. Finally, considering the imbalance in training samples between contrasts, additional data processing is required to mitigate negative impacts.

To summarize the contribution of this work, we introduce UPCMR, an unrolled model that integrates two kinds of learnable prompts into the UNet structure within each block. This design leverages the two kinds of prompts to guide the model and exploit spatio-temporal correlations across image sequences, thereby enhancing reconstruction performance and improving adaptability to various random sampling scenarios. Furthermore, we explored two variants of the curriculum learning training strategy and selected the more effective one for versatile random sampling scenarios. Nonetheless, further refinements to the training process and model design are still needed to realize the full potential of UPCMR method. 
\\
\\
\textbf{Acknowledgement.} This work is part of the project ROBUST: Trustworthy AI-based Systems for Sustainable Growth with
project number KICH3.LTP.20.006, which is (partly)
financed by the Dutch Research Council (NWO), Philips Research, and the Dutch Ministry of Economic Affairs
and Climate Policy (EZK) under the program LTP KIC
2020-2023.

\end{document}